\documentclass[letterpaper, 10 pt, conference]{IEEEtran}
\IEEEoverridecommandlockouts 

\ifdefined\pdfminorversion\pdfminorversion=4\fi

\usepackage[T1]{fontenc}

\usepackage{amsmath,amssymb,amsfonts,bm}

\usepackage{graphicx}
\usepackage{multirow}
\usepackage{array}
\usepackage{booktabs}
\usepackage{capt-of} 

\usepackage{cite}

\usepackage[ruled]{algorithm2e}
\usepackage{algorithmic}

\usepackage{textcomp}
\usepackage{xcolor}
\usepackage{gensymb}
\usepackage{outlines}
\usepackage{enumitem}

\setlist[itemize]{nosep, leftmargin=1.2em}
\makeatletter
\g@addto@macro\normalsize{%
  \setlength\abovedisplayskip{1pt plus 1pt minus 1pt}%
  \setlength\belowdisplayskip{1pt plus 1pt minus 1pt}%
  \setlength\abovedisplayshortskip{1pt}%
  \setlength\belowdisplayshortskip{1pt}}
\@ifundefined{IEEEbibitemsep}{}{\def\IEEEbibitemsep{0pt plus .5pt}}
\makeatother
\usepackage[normalem]{ulem}

\usepackage[hidelinks]{hyperref}
\hypersetup{
    pdftitle={SaddleSense: Saddle-Shaped Modules for 3D Shape Sensing with Single-Axis Off-the-Shelf Bend Sensors},
    pdfauthor={Dong Heon Han, Runze Zuo, Chankyo Kim, Saima Jamal, Daniel Bruder},
    pdfsubject={},
    pdfkeywords={}
}

\title{\LARGE \bf SaddleSense: Saddle-Shaped Modules for 3D Shape Sensing with Single-Axis Off-the-Shelf Bend Sensors}

\author{Dong Heon Han$^{1}$, Runze Zuo$^{2}$, Chankyo Kim$^{1}$, Saima Jamal$^{2}$, Daniel Bruder$^{2}$
\thanks{$^{1}$Robotics Department, University of Michigan, Ann Arbor, MI, USA.
        {\tt\small \{dongheon, chankyo\}@umich.edu}}%
\thanks{$^{2}$Department of Mechanical Engineering, University of Michigan, Ann Arbor, MI, USA.
        {\tt\small \{runzezuo, saimaj, dbruder\}@umich.edu}}%
}

\makeatletter
\let\@oldmaketitle\@maketitle
\renewcommand{\@maketitle}{\@oldmaketitle
  \vspace{0pt}
  \begin{minipage}{\textwidth}
    \centering
    \includegraphics[width=\textwidth]{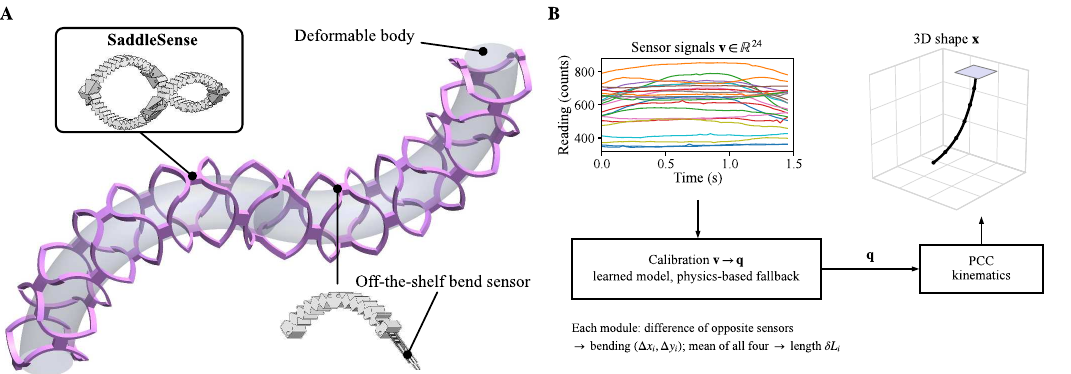}
    \setcounter{figure}{0}
    \captionof{figure}{Overview of SaddleSense. (A) Modular saddle-shaped rings are mounted along a deformable continuum body, with each curved channel holding one commercial single-axis bend sensor (four per module around the circumference). (B) A hybrid calibration---combining a learned model with a physics-based fallback for inputs far from the training distribution---maps the sensor signals $\mathbf{v}$ to piecewise constant curvature (PCC) parameters $\mathbf{q}$, from which PCC kinematics reconstructs the 3D shape $\mathbf{x}$.}
    \label{fig:fig1}
  \end{minipage}
  \vspace{2pt}}
\makeatother
\begin{document}

\maketitle
\thispagestyle{empty}
\pagestyle{empty}

\begin{abstract}
Accurate posture feedback is essential for precise control of continuum robots. Off-the-shelf bend sensors measure curvature about only one axis; without a suitable mounting arrangement, they can restrict robot motion and provide incomplete posture information. SaddleSense introduces a 3D-printed thermoplastic polyurethane (TPU) saddle-shaped lattice that holds one commercial bend sensor along each curved edge. Because each sensor flexes within its own channel rather than spanning the segment axially, the lattice converts multi-axis body deformation into sensor curvature without tying segment length to sensor length. Each four-sensor module accommodates two-axis bending, contraction, and extension while quantifying all three deformation modes across its four readings. A hybrid calibration maps the sensor readings to piecewise constant curvature (PCC) parameters using a learned model paired with a physics-based fallback for out-of-distribution inputs. We demonstrate a six-module prototype with 24 off-the-shelf bend sensors on a 430~mm continuum body. Using only each module's four local readings, the physics-based model alone achieves a tip position root mean square error (RMSE) of 29.9~mm (7.0\% of body length). The data-driven model reduces the tip RMSE to 13.2~mm (3.1\%) on held-out samples of motions similar to the training set and 15.8~mm (3.7\%) on separately recorded trials. Reducing each module from four sensors to one nearly triples the tip error.
\end{abstract}


\section{Introduction}

Continuum robots deform continuously along their bodies to navigate confined spaces \cite{webster2010design} and interact safely with their surroundings \cite{rus2015design}. However, because external loads easily alter their compliant shape \cite{majidi2014soft}, actuator inputs alone cannot uniquely determine the robot's posture \cite{wang2018toward,bruder2020data}. Closed-loop control therefore requires onboard sensors that measure multi-axis bending, contraction, and extension without restricting the robot's natural motion \cite{della2023model}.

To meet this need, we present SaddleSense (Fig.~\ref{fig:fig1}), a modular 3D-printed thermoplastic polyurethane (TPU) sleeve that houses low-cost, off-the-shelf single-axis bend sensors. Its saddle-shaped lattice converts 3D segment deformation---bending about both axes, contraction, and extension---directly into in-plane curvature changes across four sensors per module, providing sensor responses to all three deformation modes. Because the sleeve mounts externally around the robot body, it is designed for generalized continuum robot bodies, including existing systems that currently depend on external motion capture \cite{zuo2025umarm}, enabling standalone posture feedback after calibration on the target body.

External sensing systems, such as ultrasonic \cite{shi2023soft} or magnetic tracking \cite{song2018preliminary}, require dedicated infrastructure around the workspace \cite{zhou2024integrated}. Proprioceptive sensors instead measure deformation directly on or within the robot body without external tracking \cite{kramer2011soft}. Many proprioceptive approaches rely on custom-fabricated sensors, including resistive \cite{soter2018bodily,truby2019soft,muth2014embedded,truby2020distributed,so2021shape}, capacitive \cite{tairych2019capacitive,hu2022super,roberts2013soft}, fiber-optic \cite{galloway2019fiber,lun2019real}, magnetic \cite{mitchell2021fast}, and liquid-metal designs \cite{yin2020closing,shah2023stretchable,tapia2020makesense}. Compared to custom fabrication, commercial off-the-shelf sensors simplify sourcing, integration, and replacement \cite{han2025enhanced,martin2022proprioceptive}.

The most common off-the-shelf options are bend sensors and inertial measurement units (IMUs) \cite{hegde2023sensing}. Standard single-axis bend sensors support planar feedback \cite{gerboni2017feedback,han2025enhanced}, whereas commercial two-axis sensors \cite{bendlabs2019} measure angles in two planes, but neither directly measures axial length change. Combining IMUs with a piecewise constant curvature (PCC) model \cite{webster2010design} enables 3D shape estimation \cite{martin2022proprioceptive,peng2024tendon,stella2023soft}, yet orientation readings alone cannot determine segment length (Table~\ref{tab:prior}). Furthermore, when bend sensors \cite{gerboni2017feedback,han2025enhanced,zhang2023sliding} or strain sensors \cite{truby2020distributed,wang2024sensing} are mounted axially along a segment, an inextensible sensor restricts axial stretching and ties the segment length to the sensor length, while out-of-plane bending twists the sensor off its sensitive axis. Circumferential differential sensing is established in multicore fiber Bragg gratings \cite{moore2012shape}; SaddleSense realizes it mechanically with inextensible, low-cost bend sensors.

Whereas forward mechanics models map actuation or configuration to shape \cite{mbakop2021inverse,till2019real}, proprioception requires calibrating the inverse map from nonlinear, hysteretic sensor readings \cite{kim2021review} to robot configuration using mechanics-based \cite{tapia2020makesense} or learned estimators \cite{thuruthel2019soft,truby2020distributed}.
The primary contribution is a novel saddle-shaped sensing module whose curved channels enable inextensible, commercial single-axis bend sensors to capture two-axis bending and axial length changes independently of sensor length; the calibration models decode the deformation information encoded by this mechanical design.

The main contributions of this work are:
\begin{itemize}
    \item a modular saddle-shaped sleeve whose curved channels let four commercial single-axis bend sensors per segment flex in-plane, decoupling sensor length from segment length to measure two-axis bending, contraction, and extension;
    \item a PCC-based shape reconstruction pipeline that combines a learned temporal calibration with a per-module physics-based fallback for out-of-distribution states;
    \item experimental validation on a six-module, 430~mm continuum body demonstrating a tip RMSE of 7.0\% of body length using the physics-based model alone, 3.1\% on held-out trials and 3.7\% on separately recorded trials using the hybrid calibration, and a sensor-count ablation showing that reducing each module from four sensors to one nearly triples the error.
\end{itemize}
\begin{figure}
    \centering
    \includegraphics[width=\columnwidth]{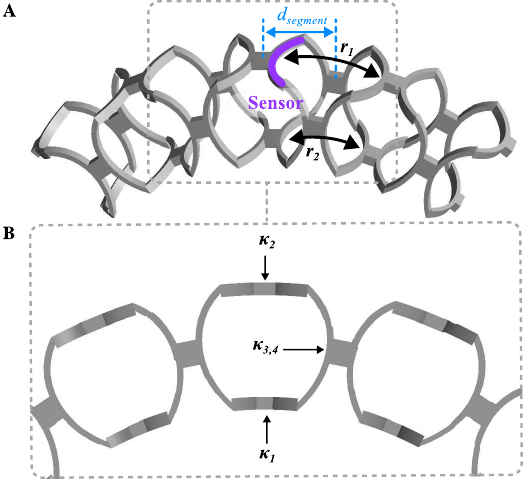}
    \caption{SaddleSense module geometry and deformation. (A) Under segment bending, each sensor (purple) flexes within its circumferential channel while the segment length $d_{segment}$ remains independent of the sensor length. Bending widens the outer channel radius ($r_1$) and tightens the inner channel radius ($r_2$). (B) Close-up of the boxed module in (A), illustrating how bending induces distinct curvature changes ($\kappa_1$ to $\kappa_{3,4}$) across the four sensors.}
    \label{fig:shape}
\end{figure}

\begin{figure*}
    \centering
    \includegraphics[width=\textwidth]{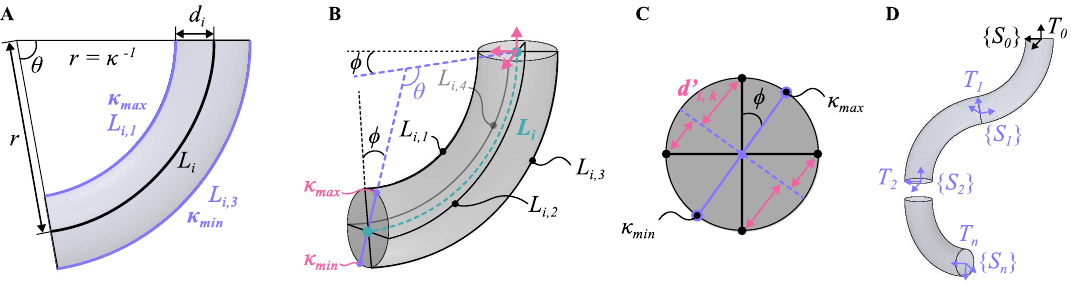}
    \caption{Piecewise constant curvature (PCC) model of one segment. (A) Planar bending with angle \(\theta\) and centerline arc length \(L_i\). The inner and outer arcs at radial offset \(d_i\) have curvatures \(\kappa_{\max}\) and \(\kappa_{\min}\). (B) Spatial bending in a plane oriented at angle \(\phi\). (C) Segment cross-section showing the four circumferential arcs \(L_{i,1}\) to \(L_{i,4}\) at projected offsets \(d'_{i,k}\) from the neutral bending axis. (D) Sequential segment transformations are chained in \(\mathrm{SE}(3)\) to reconstruct the 3D backbone.}
    \label{fig:pcc}
\end{figure*}

\section{Sensing Structure}
\label{sec:design}

\subsection{Sensing Principle}
Each SaddleSense module surrounds a robot segment with four pre-curved channels spaced 90\degree{} apart around the circumference, with each channel housing one commercial single-axis bend sensor (Fig.~\ref{fig:fig1}A, Fig.~\ref{fig:shape}). Rather than mounting sensors axially along the robot, this saddle-shaped arrangement converts 3D segment deformation---two-axis bending, contraction, and extension---directly into in-plane curvature changes across the four sensors while decoupling the segment length (\(d_{segment}\) in Fig.~\ref{fig:shape}A) from the physical sensor length. As illustrated in Fig.~\ref{fig:shape}A and B, bending compresses the inner side of the segment and stretches the outer side, which increases the curvature of the inner sensor (\(\kappa_1\)) and decreases that of the opposite outer sensor (\(\kappa_2\)) while leaving the perpendicular pair (\(\kappa_{3,4}\)) nearly unchanged. Conversely, pure axial contraction or extension changes the curvature of all four sensors together in the same direction. Thus, taking the difference between each opposite sensor pair isolates the bending component along that pair's axis (\ref{eq:config}), while taking the mean of all four readings isolates axial length change (\ref{eq:arcs}), encoding all three PCC parameters of a segment (Section~\ref{sec:kinematics}).

\subsection{Designing and Assembling the Sensing Modules}
Each sensing module consists of two rigid end connectors linked by four curved thermoplastic polyurethane (TPU) channels positioned 90\degree{} apart at a radial offset \(d_i\) from the centerline (Fig.~\ref{fig:rig}, inset~3). Adjacent modules share a rigid connector and are rotated by 90\degree{} about the body axis, forming a compliant saddle-shaped lattice that can be chained to any desired number of PCC segments.

Each TPU channel is 3D-printed with an internal pouch so that a bend sensor can be slid in from one open end, leaving only its electrical leads exposed (Fig.~\ref{fig:rig}, insets~1 and~2). To prevent out-of-plane twisting of the single-axis sensor, the channel cross-section alternates between thin flexure hinges and thick rigid ribs along its length. This corrugated profile guides the channel to flex cleanly within its own plane along the sensor's sensitive axis while holding the sensor in a pre-curved state at rest.

\section{Kinematic Model and Calibration}
\label{sec:kinematics}
\subsection{Piecewise Constant Curvature Model}
We model the continuum body under the piecewise constant curvature (PCC) assumption, where each module corresponds to one short constant-curvature segment.

For planar bending (Fig.~\ref{fig:pcc}A), the centerline curvature \(\kappa\) and the inner and outer arc curvatures \(\kappa_{\max}\) and \(\kappa_{\min}\) at radial offset \(d_i\) are
\begin{equation}
\kappa = \frac{\theta}{L}, \quad \kappa_{\max} = \frac{\theta}{L - \theta d_i}, \quad \kappa_{\min} = \frac{\theta}{L + \theta d_i},
\end{equation}
where \(\theta\) is the bending angle and \(L = L_{0,i} + \delta L_i\) is the centerline arc length, with rest length \(L_{0,i}\) and axial length change \(\delta L_i\).

When segment \(i\) bends in a plane oriented at angle \(\phi_i\) (Fig.~\ref{fig:pcc}B), the four circumferential sensor locations (where 1--3 and 2--4 form orthogonal opposite pairs) sit at projected offsets \(d'_{i,k} = \pm d_i \cos\phi_i\) for arcs 1 and 3 and \(\pm d_i \sin\phi_i\) for arcs 2 and 4 from the neutral bending axis (Fig.~\ref{fig:pcc}C). The resulting four arc lengths are
\begin{equation}
\begin{aligned}
L_{i,1/3} &= L_{0,i}+\delta L_i \mp \theta_i d_i\cos\phi_i,\\
L_{i,2/4} &= L_{0,i}+\delta L_i \mp \theta_i d_i\sin\phi_i.
\end{aligned}
\label{eq:arcs}
\end{equation}
The four sensors of module \(i\) track these arcs; \(\kappa_1\) and \(\kappa_2\) in Fig.~\ref{fig:shape}B correspond to arcs 1 and 3.

\subsection{Segment Configuration Parameterization}
To avoid the azimuth singularity at the straight pose (\(\theta_i = 0\)), we adopt the parameterization of Della Santina et al.~\cite{della2020improved}:
\begin{equation}
\begin{aligned}
\Delta x_i &= \frac{L_{i,3} - L_{i,1}}{2} = \theta_i d_i \cos\phi_i, \\
\Delta y_i &= \frac{L_{i,4} - L_{i,2}}{2} = \theta_i d_i \sin\phi_i.
\end{aligned}
\label{eq:config}
\end{equation}
Here, \(\Delta x_i\) and \(\Delta y_i\) are half-differences of opposite arc lengths, and the mean of all four arc lengths in (\ref{eq:arcs}) equals \(L_{0,i} + \delta L_i\). Each segment \(i\) is thus described by the configuration vector
\[
q_i = \begin{bmatrix} \Delta x_i & \Delta y_i & \delta L_i \end{bmatrix}^\top \in \mathbb{R}^3,
\]
and stacking all \(n\) segments yields the full robot configuration \(q \in \mathbb{R}^{3n}\).

\subsection{Forward Kinematics}
Let \(\Delta_i = \sqrt{\Delta x_i^2 + \Delta y_i^2}\), \(L_i = L_{0,i} + \delta L_i\), and \(u_i = \begin{bmatrix} -\Delta y_i & \Delta x_i & 0 \end{bmatrix}^\top\), with \(\hat{u}_i\) denoting the skew-symmetric matrix of \(u_i\). The homogeneous transformation of segment \(i\) relative to frame \(\{S_{i-1}\}\) is
\begin{equation}
T_i(q_i) =
\begin{bmatrix}
R_i(q_i) & t_i(q_i) \\
\mathbf{0}_{1 \times 3} & 1
\end{bmatrix},
\end{equation}
\begin{align}
R_i(q_i) &= I + \frac{\sin(\Delta_i/d_i)}{\Delta_i} \hat{u}_i + \frac{1 - \cos(\Delta_i/d_i)}{\Delta_i^2} \hat{u}_i^2, \\
t_i(q_i) &= \frac{L_i d_i}{\Delta_i^2}
\begin{bmatrix}
\Delta x_i (1 - \cos(\Delta_i/d_i)) \\
\Delta y_i (1 - \cos(\Delta_i/d_i)) \\
\Delta_i \sin(\Delta_i/d_i)
\end{bmatrix}.
\end{align}
Chaining these transformations gives the pose of segment \(i\) in the base frame \(\{S_0\}\) as \(T_i^0 = \prod_{j=1}^i T_j(q_j)\) (Fig.~\ref{fig:pcc}D).

\subsection{Extracting Ground-Truth Configuration from Motion Capture}
Ground-truth configuration labels for calibration are computed from motion-capture poses. Each connector frame has its \(z\)-axis aligned with the segment tangent. Given the measured poses \(T_{i-1}^0\) and \(T_i^0\) at the two ends of segment \(i\), the relative transformation \(T_i^{i-1} = (T_{i-1}^0)^{-1} T_i^0\) yields rotation \(R\) and translation \(t\), from which we recover
\begin{align}
\theta_i &= \arccos R[3,3], \\
\Delta x_i &= \frac{d_i\, \theta_i\, R[1,3]}{\sin\theta_i}, \quad
\Delta y_i = \frac{d_i\, \theta_i\, R[2,3]}{\sin\theta_i}, \\
\delta L_i &= \frac{\theta_i \|t\|}{2 \sin(\theta_i/2)} - L_{0,i}.
\end{align}

\subsection{Hybrid Physics-Based and Learned Calibration}
Because the mapping from raw sensor signals \(v\) to configuration \(q\) is affected by individual sensor tolerances, mounting variations, and mechanical cross-coupling between modules, we calibrate it against motion-capture ground truth (Fig.~\ref{fig:fig1}B). Our hybrid calibration combines a per-module physics-based model that directly encodes the sensing principle with a temporal neural network that compensates for hysteresis and inter-module coupling, blending the two via an out-of-distribution fallback weight.

\subsubsection{Physics-Based Model}
Each sensor \(k\) first passes through a monotone nonlinear calibration curve
\begin{equation}
g_k(v_k) = \alpha_{k,0} v_k + \sum_{j=1}^{8} \alpha_{k,j} \tanh\!\left(\frac{v_k - c_j}{h}\right) + \beta_k, \quad \alpha_{k,j} \ge 0,
\end{equation}
where the centers \(c_j\) are evenly spaced over \([-3, 3]\) in standardized units and the width \(h\) equals their spacing. The configuration of segment \(i\) is then modeled as an affine transformation of its four calibrated sensor outputs \(g_i \in \mathbb{R}^4\):
\begin{equation}
q_i = A_i g_i + a_i .
\end{equation}

\subsubsection{Learned Model}
To capture sensor hysteresis and mechanical coupling across adjacent modules, a gated recurrent unit (GRU) network with 64 hidden units processes the most recent 15 samples (0.4~s) of all 24 sensor channels, followed by a linear output layer that predicts \(q\). All inputs and outputs are standardized, and the network is trained to minimize the mean squared error on \(q\) using the Adam optimizer (learning rate \(10^{-3}\), batch size 256, up to 300 epochs, with early stopping after 30 epochs without improvement on a held-out 15\% random subset of training samples). We average the predictions of an ensemble of three GRUs trained with different random seeds.

\subsubsection{Out-of-Distribution Fallback Weight}
When external loads push the robot into configurations absent from the training data, we smoothly fall back to the physics-based model. Let \(\rho\) denote the Euclidean distance from the current standardized 24-channel sensor vector to its nearest neighbor in the training set (subsampled to every third frame). We compute the fallback weight
\begin{equation}
w = \min\!\left(1, \max\!\left(0, \frac{\rho - \rho_{95}}{\rho_{99.9} - \rho_{95}}\right)\right),
\label{eq:weight}
\end{equation}
where \(\rho_{95}\) and \(\rho_{99.9}\) are the 95th and 99.9th percentiles of the nearest-neighbor distance computed across the training trials via leave-one-trial-out cross-validation. After smoothing \(w\) with a causal 15-sample moving average, the final configuration estimate is
\begin{equation}
q = (1 - w)\, q_{\mathrm{learned}} + w\, q_{\mathrm{phys}} .
\end{equation}

\section{Experiments}

\subsection{Experimental Prototype and Hardware}
\begin{figure}
    \centering
    \includegraphics[width=\columnwidth]{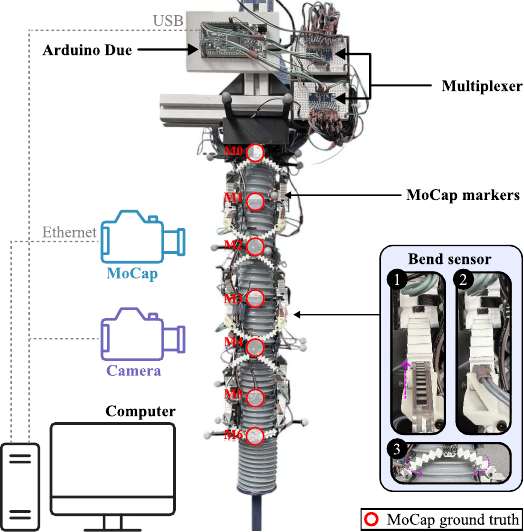}
    \caption{Experimental setup. Six SaddleSense modules cover a 430~mm flexible duct hose. An Arduino Due reads the 24 bend sensors through two analog multiplexers and streams the data to a PC over USB, while OptiTrack motion-capture markers (M0--M6) on the rigid connectors measure ground-truth segment poses. Insets: (1) sliding a bend sensor into the internal pouch of a 3D-printed TPU channel, (2) the assembled channel with only the sensor leads exposed, and (3) a channel mounted circumferentially between two connectors, with arrows indicating its in-plane flexing direction.}
    \label{fig:rig}
\end{figure}

We evaluated SaddleSense on a 430~mm flexible duct hose equipped with six sensing modules (\(n = 6\), \(q \in \mathbb{R}^{18}\)), with segment rest lengths ranging from 64 to 77~mm and radial offset \(d_i = 30\)~mm (Fig.~\ref{fig:rig}). The six modules house 24 commercial resistive bend sensors (2.2-inch Flex Sensor, Spectra Symbol; four per module), which are read at approximately 37~Hz by an Arduino Due via two analog multiplexers over USB. For ground-truth tracking, an OptiTrack motion-capture system records seven rigid marker frames (M0 to M6, mounted on the connectors bounding each segment) at 120~Hz, which we time-align via cross-correlation and interpolate onto the 37~Hz sensor time base.

\subsection{Motion and Load Trials}
Starting from the straight hanging rest pose, the continuum body was manipulated by hand for approximately 100~s per trial across three experimental conditions:
\begin{itemize}
    \item \textbf{In-distribution trials (10 trials, 36{,}727 samples):} circular and swinging motions applied at the distal tip, comprising five large-amplitude and five small-amplitude bending trials.
    \item \textbf{Separately recorded trials (3 trials, 11{,}315 samples):} free-hand tip motions following different trajectory sequences, recorded in a separate session after the in-distribution trials to test generalization.
    \item \textbf{External load trials (2 trials, 7{,}291 samples):} multi-contact deformations in which hand forces applied along the body held the hose in J, C, and S shapes, alongside localized axial compression.
\end{itemize}

\subsection{Evaluation Protocol and Metrics}
We evaluated all models using leave-one-trial-out (LOTO) cross-validation across the 13 free-motion trials (10 in-distribution and 3 separately recorded). In each fold, models were trained on 12 trials (with 15\% of the training samples held out for early stopping) and evaluated on the remaining unseen trial. For the two external-load trials, models were trained on all 13 free-motion trials. Our primary accuracy metric is the root mean square error (RMSE) of the distal tip position (M6) in the base frame (M0), reported both in millimeters and as a percentage of the 430~mm body length; we also report per-segment relative rotation errors.

\section{Results}
\label{sec:results}

\subsection{Sinusoidal Sensor Response to Bending Direction}
\begin{figure}
    \centering
    \includegraphics[width=\columnwidth]{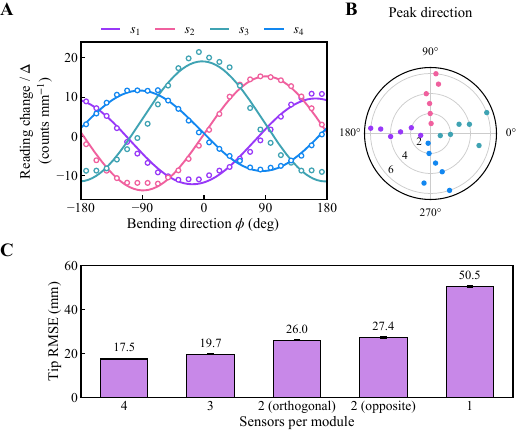}
    \caption{Validation of the module sensing principle. (A) Normalized reading change of the four sensors in segment~2 (divided by bending magnitude \(\Delta\)) versus bending direction \(\phi\) for samples with \(\Delta\) above the median (circles: binned means; curves: sinusoidal fits). (B) Peak sensitivity directions of all 24 sensors across the six segments (radial coordinate indicates segment index). (C) Sensor-count ablation showing tip position RMSE as the number of sensors per module is reduced from four to one (evaluated with an identical single-frame \(128\times 128\) MLP; error bars show standard deviation across 3 repeats).}
    \label{fig:saddle}
\end{figure}

According to the PCC geometry in (\ref{eq:arcs}), as a segment bends in direction \(\phi\), its four circumferential arc lengths vary sinusoidally with 90\degree{} phase offsets. Fig.~\ref{fig:saddle}A plots the normalized sensor reading change (divided by the segment bending magnitude \(\Delta\)) against the bending direction \(\phi\) for the four sensors of a representative module across all trials. All 24 sensors closely follow sinusoidal tuning curves (\(R^2 = 0.86\text{--}0.99\)), with the four peak sensitivity directions within each module spaced approximately 90\degree{} apart (Fig.~\ref{fig:saddle}B).

\begin{figure*}
    \centering
    \includegraphics[width=\textwidth]{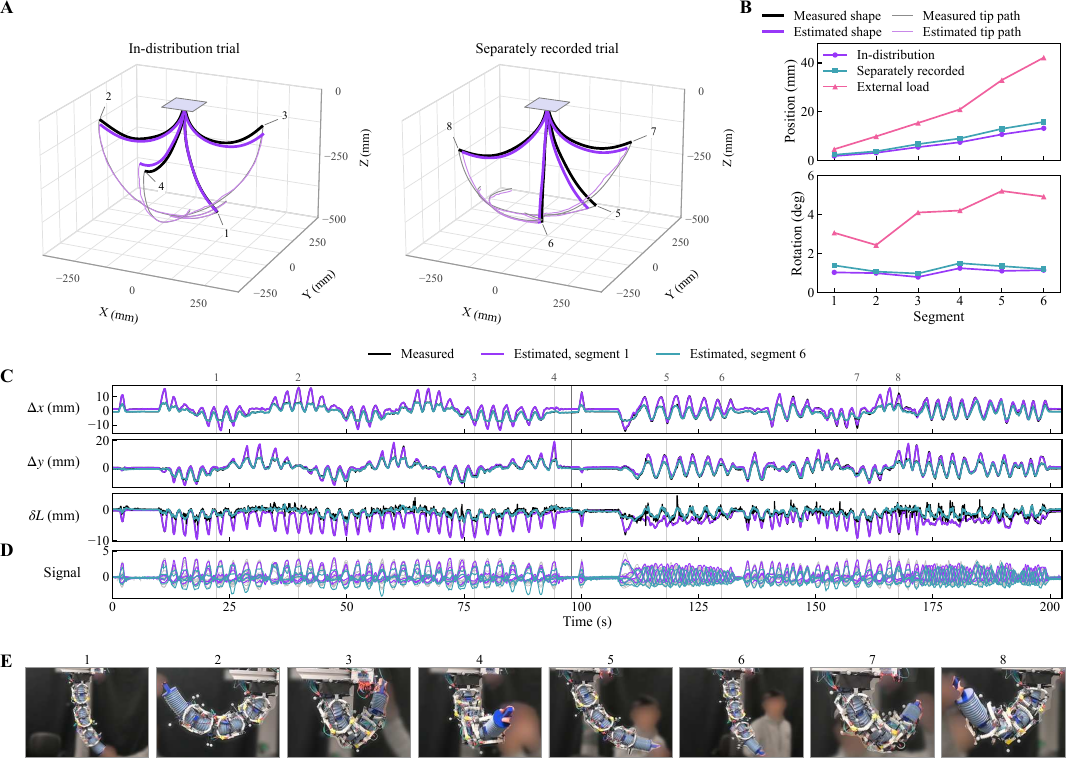}
    \caption{Continuous 3D shape reconstruction on two held-out trials (left of vertical divider: in-distribution trial; right: separately recorded trial). (A) Ground-truth (black) and reconstructed (purple) 3D backbones at eight representative large-bending instants (four per trial), along with the preceding 1.2~s tip trajectory. Labels 1--8 correspond to the vertical markers in (C)--(D) and the photographs in (E). (B) Mean position RMSE and relative rotation error across segments~1 to~6 for each trial category. (C) Ground-truth (black) and estimated PCC parameters for segment~1 (purple) and segment~6 (teal). (D) Standardized sensor readings for segment~1 (purple) and segment~6 (teal). (E) Reference camera frames at instants 1--8.}
    \label{fig:heldout}
\end{figure*}

\begin{figure*}
    \centering
    \includegraphics[width=\textwidth]{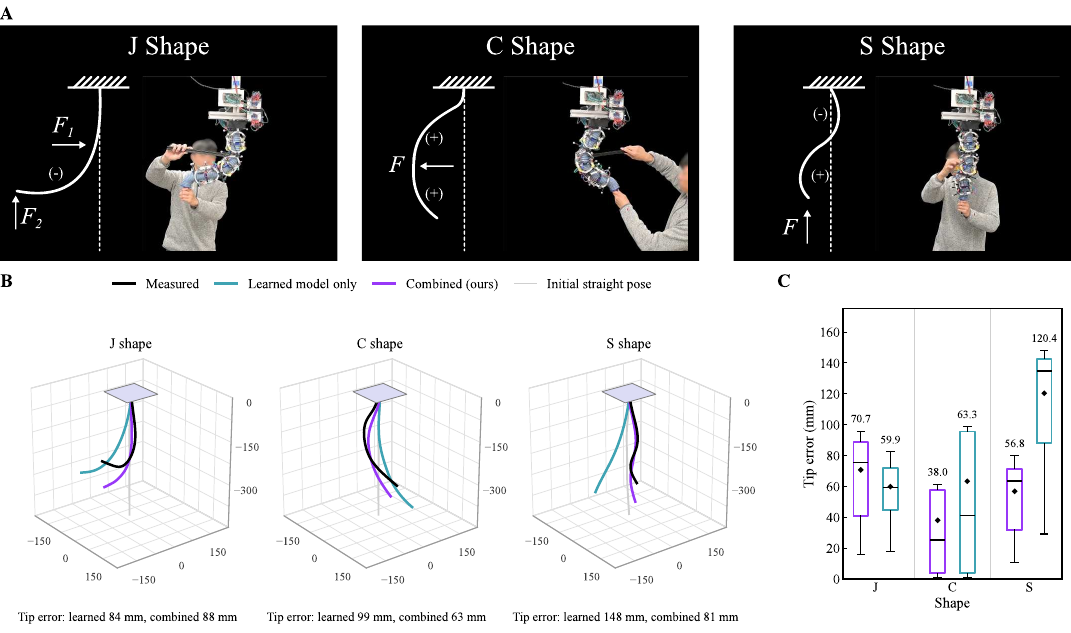}
    \caption{Shape estimation under out-of-distribution external loads. (A) Schematic of applied hand forces \(F\) and curvature signs \((+, -)\) alongside photographs of the J, C, and S load cases. (B) Ground-truth (black) and reconstructed (purple) 3D shapes at the peak-bending frame of each case (light gray: straight rest pose; tip errors listed below each plot). (C) Distribution of tip position errors across each load shape for the hybrid combined estimator (purple) versus the learned model alone (teal), showing quartiles (boxes), 5th--95th percentiles (whiskers), and RMSE (diamonds and numerical labels).}
    \label{fig:extreme}
\end{figure*}

\subsection{Decoupling Two-Axis Bending via Opposite Sensor Pairs}
\label{sec:pairs}
The peak sensitivity directions in Fig.~\ref{fig:saddle}B naturally pair opposite sensors within each module: the two sensors peaking near 0\degree{} and 180\degree{} form one differential pair aligned with \(\Delta x_i\), while those peaking near \(\pm 90\)\degree{} form the orthogonal pair aligned with \(\Delta y_i\). Across all 13 free-motion trials and all six segments, the half-difference of each opposite pair's standardized readings correlated strongly with its aligned bending coordinate (\(r = 0.97\text{--}0.98\) for \(\Delta x_i\) and \(r = 0.93\text{--}0.98\) for \(\Delta y_i\)), while exhibiting minimal cross-talk with the orthogonal bending axis (\(|r| \le 0.23\)). These results confirm that opposite sensor pairs cleanly decouple the two orthogonal bending axes of a segment.

\subsection{Ablation on the Number of Sensors per Module}
To quantify the benefit of using four sensors per module, we trained an identical single-frame \(128\times 128\) MLP using subsets of sensors in each module (Fig.~\ref{fig:saddle}C). Using four, three, two orthogonal, two opposite, and one sensor per module produced tip RMSEs of 17.5, 19.7, 26.0, 27.4, and 50.5~mm, respectively. While two orthogonal sensors span both bending axes and two opposite sensors capture one bending axis plus axial length, a single sensor per module cannot distinguish bending direction from bending magnitude, nearly tripling the tip error compared to the full four-sensor module.

\subsection{Sensing Axial Contraction and Extension}
\label{sec:axial}
Across the in-distribution and separately recorded trials, the combined estimator tracked axial length changes \(\delta L \in [-9.5, 3.6]\)~mm (0.1st to 99.9th percentile) with an RMSE of 0.37 and 0.46~mm, respectively, and a median correlation of \(r = 0.98\) (Fig.~\ref{fig:heldout}C). To evaluate pure axial deformation without bending, we examined segments subjected to direct axial compression (\(\delta L < -8\)~mm) while remaining nearly straight (\(\theta_i < 5\degree\)) during the S-shape load trial. Across segments 1, 3, 4, and 6, 731 segment-frames met this pure-compression criterion, with the largest compression occurring in distal segment~6 (461 frames, reaching \(\delta L = -27.5\)~mm), well beyond the training distribution (\(\delta L \ge -11.9\)~mm, always accompanied by bending). As predicted by our sensing principle, all four sensors in the compressed module increased together: their mean standardized reading rose by 1.4 to 2.6 standard deviations (compared to \(-0.17\) at rest and \(0.30\) during bending above 15\degree{}), while opposite-pair differences stayed below 0.7. Across these compressed frames and the corresponding rest frames, the module mean reading correlated strongly with \(\delta L\) (\(r = -0.77\) to \(-0.94\)).

\subsection{3D Shape Reconstruction Accuracy}
Fig.~\ref{fig:heldout} illustrates continuous 3D shape reconstruction on two held-out trials. Our combined estimator achieved a tip position RMSE of 13.2~mm (3.1\% of body length) across the 10 in-distribution trials and 15.8~mm (3.7\%) across the 3 separately recorded trials, with mean relative rotation errors of 1.1\degree{} and 1.2\degree{} per segment. While cumulative position error grows from the base toward the distal tip, the angular error per segment remains uniform along the body (Fig.~\ref{fig:heldout}B).

\subsection{Comparison of Calibration Models}
Table~\ref{tab:models} compares calibration architectures evaluated on the same dataset. Applying a single global linear map across the 24 raw sensor readings achieved a tip RMSE of 33.0~mm, and our per-module physics-based model---an affine map of each module's four local calibrated sensors using only 330 parameters in total---lowered it further to 29.9~mm (7.0\% of body length), demonstrating that the SaddleSense geometry captures the majority of the 3D shape directly in the local readings. Incorporating temporal context across all 24 channels via learned models (spanning 5 frames / 0.13~s for the MLP ensemble or 15 frames / 0.4~s for the GRU ensemble) reduced the in-distribution tip RMSE to 13.2--14.1~mm, with the GRU ensemble performing best.

\begin{table}
\caption{Tip position RMSE (mm) across calibration models. In-dist.\ and Sep.\ rec.\ report LOTO cross-validation over 13 trials (mean of 3 repeats). Ext.\ load models are trained on all 13 free-motion trials (mean of 3 repeats; --: not evaluated). Linear: global linear map of the 24 readings. MLP: \(128\times 128\) units over 5 frames. GRU: 64 units over 15 frames. Ens.: ensemble mean of 3 networks. Combined: GRU ensemble blended with the physics-based fallback.}
\label{tab:models}
\centering
\small
\setlength{\tabcolsep}{3pt}
\begin{tabular}{lrrrr}
\toprule
Model & Params & In-dist. & Sep. rec. & Ext. load \\
\midrule
Linear & 450 & 33.0 & 32.9 & -- \\
Physics-based & 330 & 29.9 & 30.0 & 43.6 \\
MLP ens. & 103k & 14.1 & 18.2 & 83.8 \\
GRU ens. & 55k & 13.2 & 16.1 & 66.3 \\
Combined (ours) & 56k & 13.2 & 15.8 & 42.3 \\
\bottomrule
\end{tabular}
\end{table}

\subsection{Generalization Under External Loads}
Under hand-applied external loads (Fig.~\ref{fig:extreme}), the learned GRU ensemble alone degraded to a tip RMSE of 66.3~mm (15.4\% of body length) as multi-point contacts forced the body into configurations unseen during free-motion training. In these regimes, the out-of-distribution weight \(w\) automatically shifted toward the physics-based model, averaging 0.51 and 0.78 across the two load trials compared to 0.00--0.11 during free-motion trials. Blending in the physics-based fallback reduced the overall load-trial tip RMSE to 42.3~mm (9.8\%), matching the 43.6~mm accuracy of the standalone physics-based model (Table~\ref{tab:models}). Specifically, the fallback substantially reduced tip error in the C-shape (from 63.3 to 38.0~mm) and S-shape (from 120.4 to 56.8~mm) load cases, though error increased in the J-shape case (from 59.9 to 70.7~mm).

\section{Discussion and Limitations}
Notably, the physical SaddleSense arrangement itself encodes most of the posture information: a lightweight physics-based model with only 330 parameters that maps each module's four local sensors independently already achieves a tip RMSE of 29.9~mm (7.0\% of body length). Adding the temporal learned model compensates for sensor hysteresis and inter-module coupling, cutting the error in half on motions similar to the training set (3.1\%--3.7\%). Conversely, under out-of-distribution external loads where purely data-driven networks extrapolate poorly, our distance-based fallback mechanism recovers the robustness of the decoupled per-module physics-based model under C- and S-shape external loads (though not in the J-shape case).

Among the commercial-sensor systems in Table~\ref{tab:prior}, SaddleSense uniquely recovers both 3D bending and axial length change using only single-axis bend sensors.

\begin{table}[t]
\caption{Comparison of continuum shape-estimation methods
using embedded sensors. Tip errors are reported by each
study; robot bodies, motion regimes, and evaluation
metrics differ across works.}
\label{tab:prior}
\centering
\scriptsize
\setlength{\tabcolsep}{3pt}
\begin{tabular}{@{}llccl@{}}
\toprule
Work & Sensors & 3D & $\delta L$ & Tip error \\
\midrule
Martin et al.~\cite{martin2022proprioceptive}
& IMUs & $\checkmark$ & $\times$ & 6\% of length \\
Peng et al.~\cite{peng2024tendon}
& IMUs & $\checkmark$ & $\times$ & 20~mm \\
Stella et al.~\cite{stella2023soft}
& IMUs & $\checkmark$ & $\times$ & 7\% of length \\
Han et al.~\cite{han2025enhanced}
& IMUs, bend & $\times$ & $\times$ & 2.91\% of length \\
\textbf{SaddleSense}
& Bend & $\checkmark$ & $\checkmark$ & 3.1\% / 3.7\% of length$^\ast$ \\
\bottomrule
\end{tabular}

\smallskip
\parbox{\columnwidth}{\footnotesize
3D: two-axis bending; $\delta L$: axial length change.
$^\ast$Tip position RMSE on held-out samples and
separately recorded trials, respectively.}
\end{table}

Several limitations remain for future investigation. First, SaddleSense was tested on a passive continuum hose manipulated by hand rather than an actively driven robot, and we did not benchmark against an axial sensor mounting on the same body or quantify the added mass and stiffness of the TPU sleeve. Second, axial length changes \(\delta L\) in the training set occurred only in conjunction with bending; pure axial compression was encountered only in an external-load trial and was underestimated once compression exceeded the training range. Third, fitting the calibration model requires ground-truth motion capture of the target body, and long-term sensor hysteresis and drift over hours of operation remain to be evaluated beyond our 100~s trials.

\section{Conclusion}
We presented SaddleSense, a modular proprioceptive sensing sleeve that arranges four commercial single-axis bend sensors in pre-curved TPU channels around each segment of a continuum body. By converting 3D body deformations into in-plane sensor curvature, opposite sensor pairs differentially measure two-axis bending while the four-sensor mean captures axial contraction and extension, decoupling segment length from sensor length. Across a six-module, 430~mm prototype, individual sensors exhibited sinusoidal directional tuning spaced 90\degree{} apart, and our hybrid physics-informed neural calibration achieved tip position RMSEs of 3.1\% of body length on held-out in-distribution motions and 3.7\% on separately recorded trajectories. 

Future work will adapt the modular sleeve to actuated continuum robots with different diameters, segment lengths, and mechanical properties to assess its applicability beyond the present hose prototype. These studies will quantify the sleeve's added stiffness, evaluate sensing accuracy under independent axial and bending deformations, and determine the calibration required for each robot. We will further investigate long-term sensor drift and demonstrate closed-loop control using SaddleSense feedback under external loading.

\section*{ACKNOWLEDGMENT}
Claude Code was used to assist with code development, brainstorming, and language refinement.

\bibliographystyle{IEEEtran}
\bibliography{main}

\end{document}